\documentclass[conference]{IEEEtran}
\IEEEoverridecommandlockouts

\usepackage{cite}
\usepackage{amsmath,amssymb,amsfonts}
\usepackage{algorithmic}
\usepackage{graphicx}
\usepackage{textcomp}
\usepackage{algorithm}
\usepackage{algorithmic}
\usepackage{multirow}
\usepackage[hyphens]{url}  
\usepackage{graphicx} 
\usepackage{multirow}
\usepackage[table]{xcolor}
\usepackage{hyperref}
\usepackage{newfloat}
\usepackage{listings}

\def\BibTeX{{\rm B\kern-.05em{\sc i\kern-.025em b}\kern-.08em
    T\kern-.1667em\lower.7ex\hbox{E}\kern-.125emX}}
\begin{document}

\title{MILES: Modality-Informed Learning Rate Scheduler for Balancing Multimodal Learning
}

\author{\IEEEauthorblockN{Alejandro Guerra-Manzanares and Farah E. Shamout}
\IEEEauthorblockA{Division of Engineering\\ New York University Abu Dhabi, Abu Dhabi, UAE\\
Email: \{alejandro.guerra, farah.shamout\}@nyu.edu}}


\maketitle

\begin{abstract}
The aim of multimodal neural networks is to combine diverse data sources, referred to as modalities, to achieve enhanced performance compared to relying on a single modality. However, training of multimodal networks is typically hindered by modality overfitting, where the network relies excessively on one of the available modalities. This often yields sub-optimal performance, hindering the potential of multimodal learning and resulting in marginal improvements relative to unimodal models. In this work, we present the Modality-Informed Learning ratE Scheduler (MILES) for training multimodal joint fusion models in a balanced manner. MILES leverages the differences in modality-wise conditional utilization rates during training to effectively balance multimodal learning. The learning rate is dynamically adjusted during training to balance the speed of learning from each modality by the multimodal model, aiming for enhanced performance in both multimodal and unimodal predictions. We extensively evaluate MILES on four multimodal joint fusion tasks and compare its performance to seven state-of-the-art baselines. Our results show that MILES outperforms all baselines across all tasks and fusion methods considered in our study, effectively balancing modality usage during training. This results in improved multimodal performance and stronger modality encoders, which can be leveraged when dealing with unimodal samples or absent modalities. Overall, our work highlights the impact of balancing multimodal learning on improving model performance. 
\end{abstract}


\begin{IEEEkeywords}
multimodal learning, modality overfitting, learning rate scheduler, balanced training, joint fusion networks
\end{IEEEkeywords}

\section{Introduction}

Real-world prediction tasks are often multimodal in nature. From everyday activities like interpersonal communication \cite{levinson2014origin}, learning and cognition \cite{ward2017enhanced}, to specialized tasks such as stock investment \cite{galeotti1994stock} and clinical differential diagnosis \cite{langlois2002making}, humans integrate multiple sources of information to make accurate decisions and predictions. Despite this, most available datasets and machine learning models are unimodal, such that they rely on a single source of information, or modality, for a given prediction task. This dominant modeling approach overlooks the potential performance enhancements that could be achieved by combining multiple input modalities. 

Multimodal machine learning seeks to enhance model performance by integrating related information from multiple sources. While several multimodal learning approaches have been successfully developed for specific problems \cite{poria2017review, baltruvsaitis2018multimodal}, multimodal training presents inherent challenges. These challenges include the need for effective data fusion, processing heterogeneous data modalities, and ensuring that the model can leverage the complementary nature of the multimodal data without being hindered by noise or inconsistencies. Furthermore, the alignment and synchronization of disparate data sources can be complex, especially when dealing with heterogeneous datasets that vary in format, scale, and quality.

\begin{figure}[t]
\centering
\includegraphics[width=0.97\columnwidth]{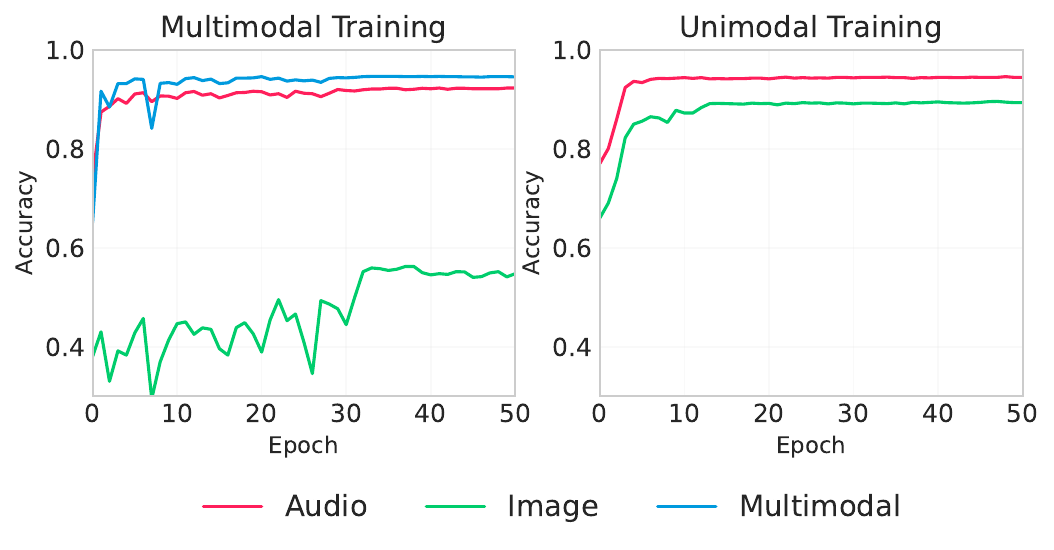}
\caption{Validation accuracy of multimodal model and its unimodal encoders (left) and unimodal models trained independently (right) using the S-MNIST dataset.}
\label{fig:overfit}
\end{figure}

Another significant challenge in multimodal machine learning is balancing the contributions of each modality to the learning process. Naive multimodal training leads to modality competition, where one modality becomes more dominant \cite{wu2022characterizing}. The multimodal model would then overfit to the dominant modality while underutilizing the remaining available modalities \cite{wu2022characterizing}. As an example, Fig. \ref{fig:overfit} shows naive runs for unimodal and multimodal models trained on the S-MNIST dataset. The left figure shows the training accuracy of the unimodal and multimodal encoders within a multimodal model, while the right figure shows validation accuracy of unimodal models trained independently. We can observe that while the image model has potential to reach a similar accuracy to that of the audio (right figure), it does not reach its full potential in the multimodal model (left figure), showing significantly lower performance. On the other hand, the audio encoder achieves comparable performance in both the unimodal and multimodal settings. The multimodal model (in blue) also tends to overfit to the audio model. Over-reliance on one modality may lead to sub-optimal performance, especially if that modality is noisy or less informative in certain contexts. Conversely, underutilization of a critical modality could result in missed opportunities to improve prediction accuracy. Additionally, the computational cost and model complexity tend to increase with the number of modalities considered, posing further obstacles to efficient model training and deployment.

To address the challenge of training multimodal networks in a balanced manner, we propose the Modality-Informed Learning ratE Scheduler (MILES). MILES is a learning rate scheduler for multimodal networks that leverages the conditional utilization rate per modality during training to dynamically adjust the learning rate, effectively balancing multimodal learning, increasing the contribution of the underutilized modalities and enhancing multimodal prediction performance. The main contributions of this work are summarized below:

\begin{itemize}
    \item We present the Modality-Informed Learning RatE Scheduler, MILES, a novel learning rate scheduler for multimodal training that dynamically adjusts modality-specific learning rates based on each modality's contribution during training, effectively balancing multimodal learning.
    \item We demonstrate that by increasing the utilization of non-dominant modalities and preventing overfitting to the dominant modality, MILES not only produces better multimodal results but also strengthens encoders performance, achieving accuracy comparable to unimodal models.
    \item We conduct extensive validation of MILES on four multimodal datasets using two fusion methods, demonstrating its superior performance compared to seven state-of-the-art approaches.
\end{itemize}

\section{Related Work}

\subsection{Multimodal joint fusion}

Data fusion involves combining information from multiple sources (modalities) to extract complementary insights, leading to more comprehensive and better-performing machine learning models than those relying on a single data source. The three primary fusion strategies are early fusion, joint fusion, and late fusion \cite{huang2020fusion}. Early fusion, also known as feature-level fusion, combines the available modalities into a single feature vector, which is used as input for a machine learning model. On the other hand, late fusion, also known as decision-level fusion, aggregates predictions from multiple models to compute the final prediction. Finally, joint fusion, also known as intermediate fusion, involves combining learned feature representations from intermediate layers of neural networks (encoders), for further processing. Joint fusion models are trained end-to-end, allowing feature representations to improve with each iteration. Our work focuses on multimodal learning in joint fusion networks, which is the most common setting of multimodal fusion. These networks combine features at different levels of the model, leading to a more comprehensive understanding of the data and generally better overall performance. We demonstrate the impact of our proposed approach for the most widely used fusion methods: feature concatenation and feature summation \cite{huang2020fusion}.

\subsection{Modality competition}

Even though multimodal fusion models have the potential to outperform unimodal models, the joint training of multimodal networks often fails \cite{huang2022modality} or produces a sub-optimal model that does not surpass the best unimodal model \cite{wang2020makes}. During joint training, available modalities compete with each other, leading to a situation where only a small subset of modalities are effectively learned while the others are neglected and remain underexplored \cite{huang2022modality}. In practice, as shown in previous work, only one of the available modalities dominates the learning process \cite{wu2022characterizing}. Possible reasons behind the tendency of multimodal models to overfit to one modality while neglecting others are better alignment of certain modalities with the encoding network’s random initialization \cite{huang2022modality}, faster learning speeds of specific modalities \cite{wu2022characterizing}, and differences in convergence rates and generalization performance among modalities \cite{fujimori2020modality, wang2020makes}.

\subsection{Balanced multimodal learning}

To address modality competition, avoid overfitting, and balance modality contributions, several methods have been proposed for end-to-end training of multimodal fusion networks. \cite{fujimori2020modality} introduced the Modality-Specific Early Stopping (MSES) method, which tackles the differing convergence rates and generalization performance of various modalities by applying a global learning rate to the network and using early stopping during joint training for modalities that have converged based on validation performance. The training continues until all parts of the network have converged.

Building on similar reasoning, \cite{yao2022modality} proposed the Modality-Specific Learning Rate (MSLR) method. MSLR is based on the observation that while a global learning rate may work well for some modalities (and lead to overfitting), it may be outside the effective range for others (causing them to be ignored). To address this, MSLR advocates using distinct learning rates for the multimodal model and each encoder. The proposed MSLR variants include: MSLR-K (the \emph{keep} strategy, which maintains the best fine-tuned unimodal learning rates for different modalities fixed during multimodal model training), MSLR-S (the \emph{smooth} strategy, which adjusts the learning rates of different modalities to be closer to the average learning rate across modalities), and MSLR-D (the \emph{dynamic} strategy, which scales the learning rate for each unimodal encoder during training by a small factor based on the unimodal average prediction performance on the validation set over the last several epochs).

\cite{wang2020makes} proposed Gradient Blending (G-Blend), which directly modifies the gradient descent process by altering the loss function to a weighted sum of multiple unimodal losses. The weight for each modality is computed based on an Overfitting-to-Generalization Ratio (OGR), which describes the overfitting conditions for each modality. To compute OGR, each unimodal model is trained individually for the first several epochs, using the same learning rate for both the modality-specific and multimodal models. \cite{peng2022balanced} introduced the On-the-fly Gradient Modulation (OGM) method to adaptively control the optimization of each modality, via monitoring the discrepancy of their contribution towards the
learning objective. OGM-GE refines this method by incorporating dynamic Gaussian noise during training to prevent a generalization drop from gradient modulation. 

Compared to existing approaches, MILES computes the conditional utilization rate per modality after each epoch as a proxy for each modality's contribution to the multimodal model and adjusts the learning rate only for the dominant modality accordingly. This simple yet effective adjustment does not require calculating any weights, performing uni-modal pretraining, or setting individual learning rates, which necessitate increased computational resources and hyper-parameter tuning rounds for offline unimodal model pretraining prior multimodal training. MILES only requires a global learning rate, as the modified learning rate is derived from the global learning rate using a scaling factor. All of the aforementioned state-of-the-art approaches are used as baselines in our work.

\section{Methodology}

In this section, we begin by introducing some preliminaries and the conditional utilization rate, a key component leveraged by our approach to dynamically adjust the learning rate during training. Following that, we present our proposed method, the Modality-Informed Learning ratE Scheduler (MILES), designed to address modality competition and balance multimodal utilization during the training process. 

\subsection{Preliminaries}

In this work, we assume the presence of two modalities, $ A $ and $ B $, with the goal of predicting a target variable $ \mathbf{y} $. Each modality is processed by its own encoder within a multimodal network. Let $ \mathbf{A} \in \mathbb{R}^{n \times d_A} $ and $ \mathbf{B} \in \mathbb{R}^{n \times d_B} $ represent the feature matrices for modalities $ A $ and $ B $, respectively, where $ n $ is the number of samples, $ d_A $ is the dimensionality of modality $ A $, and $ d_B $ is the dimensionality of modality $ B $. The multimodal network consists of two modality encoders, $f_A $ and $ f_B $, which transform the input features into encoded representations:

\begin{equation}
\mathbf{z}_A = f_A(\mathbf{A}), \quad \mathbf{z}_B = f_B(\mathbf{B}),
\end{equation}

where $ \mathbf{z}_A $ and $ \mathbf{z}_B $ are the latent representations of modalities $ A $ and $ B $, respectively. The goal is to predict the target variable $ \mathbf{y} $ using a function $ g $ that combines the encoded features:

\begin{equation}
\label{multimodal_loss}
\hat{\mathbf{y}}_{AB} = g_{AB}(\mathbf{z}_A, \mathbf{z}_B),
\end{equation}

where $\hat{\mathbf{y}}_{AB}$ is the prediction using the multimodal network. The prediction is learned by optimizing the network parameters of $f_A $, $f_B $ and $g_{AB} $ jointly to minimize the loss function:

\begin{equation}
    \mathcal{L} = \mathcal{L}_{AB}(\hat{\mathbf{y}}_{AB}, \mathbf{y}) + 
    \mathcal{L}_{{A}}(\hat{\mathbf{y}}_{A}, \mathbf{y}) + 
    \mathcal{L}_{{B}}(\hat{\mathbf{y}}_{B}, \mathbf{y}),
    \label{mml-loss}
\end{equation}

where $\mathcal{L}_{AB}(\hat{\mathbf{y}}_{AB}, \mathbf{y})$, is the loss for the multimodal model, $\mathcal{L}_{{A}}(\hat{\mathbf{y}}_{A}, \mathbf{y})$ refers to the loss of the encoder of modality A, $\mathcal{L}_{{B}}(\hat{\mathbf{y}}_{B}, \mathbf{y})$ for the modality B encoder. $\hat{\mathbf{y}_{i}}$ refers to the predicted label by the corresponding model and $ \mathbf{y} $ is the true label. 

\subsection{Motivation}

The \emph{greedy learner hypothesis}, introduced by \cite{wu2022characterizing}, establishes that a multimodal learning process where the network is trained to minimize modality-specific losses is inherently greedy. This process tends to rely on only one of the available input modalities — the one that is the fastest to learn from. However, as shown in Figure \ref{fig:overfit} and in related work such as by \cite{wang2020makes}, this phenomenon is also observed in joint fusion networks, which are typically trained to minimize a single multimodal loss (the $\mathcal{L}_{AB}(\hat{\mathbf{y}}_{AB}, \mathbf{y})$ term in Equation \ref{mml-loss})  or a sum of multimodal and modality-specific losses (Equation \ref{mml-loss}). To address this problem in joint fusion networks, we propose MILES, a method designed to regulate the learning speed at which modalities are learned by the multimodal model. MILES dynamically adjusts the modality-specific learning rates based on an epoch-wise estimation of each modality’s marginal contribution to the overall multimodal performance. This adjustment slows down the learning of the dominant modality, allowing the non-dominant modalities to be learned more effectively.

\subsection{Conditional utilization rate}

We redefine and repurpose the \emph{conditional utilization rate}, originally presented by\cite{wu2022characterizing} and modified by \cite{guerra-manzanares2025mind}, denoted as $\mathbf{u}$, to characterize epoch-wise modality usage in multimodal joint fusion networks. We define \textbf{u} for multimodal joint fusion networks as follows:

\begin{equation}
\label{eq:utilization}
    \mathbf{u_A} = \frac{M(\hat{\mathbf{y}}_{AB}) - M(\hat{\mathbf{y}}_{B})}{M(\hat{\mathbf{y}}_{AB})},
    \mathbf{u_B} = \frac{M(\hat{\mathbf{y}}_{AB}) - M(\hat{\mathbf{y}}_{A})}{M(\hat{\mathbf{y}}_{AB})},
\end{equation}

where $M(\cdot)$ represents the performance metric used (e.g., accuracy, F1 score), $\mathbf{u_A}$ determines the conditional utilization rate for modality $ A $, and $\mathbf{u_B}$ for modality $ B $. Intuitively, it quantifies the marginal contribution of a specific modality to the fusion model's performance.

We define $\delta_{AB}$ as the absolute value of the difference between conditional utilization rates: 

\begin{equation}
\label{eq:difference}
\delta_{AB} = | \mathbf{u_A} - \mathbf{u_B} |. 
\end{equation}

It enables the evaluation of imbalanced modality usage within the multimodal fusion model. Note that $\delta_{AB} \in \mathbb{R} : s.t.\: 0 \leq \delta_{AB} \leq 1$, with values closer to one indicating imbalanced modality usage and modality overfit, since the difference in conditional utilization rates would be high indicating that one modality is being used more than the other.

\subsection{Modality-Informed Learning ratE Scheduler (MILES)}

Our proposed methodology for addressing modality competition and enhancing multimodal training, is described in Algorithm \ref{alg:mils}. MILES only requires a global learning rate to be specified, and it re-adjusts the learning rates of individual modalities by scaling them with a factor of $\mu$ when certain conditions are met. These conditions are governed by the target difference threshold, denoted as $\tau$. By combining these two hyper-parameters ($\tau$ and $\mu$) following the conditional statements and computations defined in Algorithm \ref{alg:mils}, MILES dynamically adjusts the learning rate per modality ($\alpha_{A}$ and $\alpha_B$) during training based on differences in conditional utilization rates.



\begin{algorithm}[t!]
\small
\caption{MILES}
\label{alg:mils}
\textbf{Input}: Target difference threshold: $\tau$, Training epochs: $N$, Reduction factor: $\mu$, Global learning rate: $\alpha$ 
\begin{algorithmic}[1]
\STATE Initialize variables: $\delta_{AB} \gets 0$, $\alpha_{AB} \gets \alpha$,  $\alpha_{A} \gets \alpha$,  $\alpha_{B} \gets \alpha$
\FOR {$i = 1, \dots, N$}
\STATE  Train\_epoch($i$)
\STATE  Validation\_epoch($i$)
\item[]
\STATE Compute $\mathbf{u_A}$ \COMMENT{$\mathbf{u}$ of modality A as in Eq.~\ref{eq:utilization}}
\STATE Compute $\mathbf{u_B}$ \COMMENT{$\mathbf{u}$ of modality B as in Eq.~\ref{eq:utilization}}
\STATE Compute $\delta_{AB}$ \COMMENT{Difference between modalities as in Eq.~\ref{eq:difference}}
\item[]
\IF{$(\delta_{AB} \leq \tau ) \lor (\mathbf{u_A} < 0 \land \mathbf{u_B} < 0)$}
    \STATE $\alpha_{A} \gets \alpha_{AB}$ 
    \STATE $\alpha_{B} \gets \alpha_{AB}$  
\ELSE   
\IF{$\mathbf{u_A} > 0 \land \mathbf{u_B} < 0$}
\STATE $\alpha_{A} \gets \mu \cdot \alpha_{AB}$ 
\ELSIF{$\mathbf{u_A} < 0 \land \mathbf{u_B} > 0$}
\STATE $\alpha_{B} \gets \mu \cdot \alpha_{AB}$
\ELSE 
\IF{$\mathbf{u_A} < \mathbf{u_B}$}
\STATE $\alpha_{B} \gets \mu \cdot \alpha_{AB}$ 
\ELSE    
\STATE $\alpha_{A} \gets \mu \cdot \alpha_{AB}$ 
\ENDIF
\ENDIF
\ENDIF
\ENDFOR
\end{algorithmic}
\end{algorithm}

After every training epoch and subsequent validation stage, MILES is applied to tune the learning rates per modality for the next training iteration (lines 3-4, Algorithm \ref{alg:mils}). The first step of the MILES procedure involves calculating the conditional utilizations per modality ($\mathbf{u_{A,B}}$) using Eq. \ref{eq:utilization}, and the difference between them ($\delta_{AB}$) using Eq. \ref{eq:difference} (lines 5-7, Algorithm \ref{alg:mils}). Based on computed values $\mathbf{u_{A,B}}$, $\delta_{A,B}$ and set hyper-parameters $\tau$ (target difference threshold) and $\mu$ (reduction factor), the following conditions apply (lines 8-24):
\begin{itemize}
    \item If $\delta_{AB}$ is below or equal to the target threshold $\tau$, or $\mathbf{u_{A}} < 0$ and $\mathbf{u_{B}} < 0$, MILES takes no action. In the former case, the target threshold is met so no action from MILES is required. In the latter case, the unimodal encoders outperform the multimodal model, which typically occurs during the first few epochs of training when the multimodal model has not yet started learning effectively. In such instances, it is preferable to wait until the multimodal model begins to learn, so MILES takes no action. In either case, the learning rates for both modalities rate are set as the global learning rate for the next training epoch (lines 8-10, Algorithm \ref{alg:mils}). 
    \item If the previous condition is not satisfied, meaning $\delta_{AB} > \tau$, the following conditions and actions apply:
    \begin{itemize}
        \item If $\mathbf{u_{A}} > 0$, $\mathbf{u_{B}} < 0$, and $\delta_{AB} > 0$ then this implies that modality B is under-utilized and the learning rate of modality A is scaled by $\mu$ for the next epoch (lines 12-13). The aim is to slow down the learning of modality A, the over-utilized modality, allowing the model to learn from modality B more effectively.
        \item If $\mathbf{u_{A}} < 0$, $\mathbf{u_{B}} > 0$, and $\delta_{AB} > 0$ then this implies that modality A is under-utilized and the learning rate of modality B is scaled by $\mu$ for the next epoch (lines 14-15). This is the inverse of the previous condition, where the same rationale applies.
        \item If $\mathbf{u_{A}} > 0$ and $\mathbf{u_{B}} > 0$, then this indicates that the multimodal model is outperforming both modality encoders, suggesting that the multimodal model is learning effectively. However, to assess if there is modality overfitting, $\mathbf{u_{A}}$ and $\mathbf{u_{B}}$ need to be further compared:
        \begin{itemize}
            \item If $\mathbf{u_{A}} < \mathbf{u_{B}}$ then modality A is under-utilized and the learning rate of modality B is scaled by $\mu$ for the next epoch (lines 17-18). 
            \item If the opposite is true, $\mathbf{u_{A}} > \mathbf{u_{B}}$, then modality B is under-utilized and the learning rate of modality A is scaled by $\mu$ for the next epoch (lines 19-20). 
        \end{itemize}
        
    \end{itemize}
\end{itemize}

Note that the following assumptions also apply:

\begin{itemize}
\item $\tau \in [0.0, 1.0]$ because $\tau$ only takes possible values of $\delta_{AB}$, and the range of $\delta_{AB}$ (Eq.~\ref{eq:difference}) lies within the interval $[0.0, 1.0]$.
\item $\mu \in \left\{\frac{a}{b} \,\middle|\, a = 1, b \in \mathbb{R}, b \neq 0 \right\}$ as the usage of $\mu$ in Algorithm~\ref{alg:mils} involves multiplying it by the current learning rate to effectively reduce it, $\mu$ cannot be greater than 1.

\item An auxiliary supervised loss term is added for each encoder to the objective function (e.g., for modality A, a binary cross entropy loss term,  $\mathcal{L}_{{A}}(\mathbf{y}, \hat{\mathbf{y}}_{A})$, would be added, where $\mathbf{y}$ refers to the ground-truth labels and $\hat{\mathbf{y}}_{A}$ are the unimodal encoder predictions). 
\item Only one modality's learning rate is adjusted per epoch.
\end{itemize}

\section{Experiments}

In this section, we provide an overview of the datasets, baselines and experimental settings used in the main experiments and ablation studies. We utilize various multimodal datasets, used as benchmarks in prior work, either in their unimodal or multimodal versions. For reproducibility, we make our implementation publicly available at \href{https://github.com/nyuad-cai/MILES}{https://github.com/nyuad-cai/MILES}.

\subsection{Datasets}

\textbf{CREMA-D} \cite{cao2014crema} is an audio-visual dataset for multimodal emotion recognition. It includes 7,442 video clips from 91 actors speaking a selection of 12 sentences. The utterances express one of six common emotions: anger, happiness, disgust, fear, neutral, and sadness. We split the full dataset into training, validation, and test sets with a 70-15-15 split. The training set includes 5,210 samples, while both the validation and test sets include 1,116 samples.


\textbf{S-MNIST} \cite{smnistdataset} is an audio-visual dataset designed for benchmarking multimodal classification. It pairs the original MNIST \cite{lecun1998mnist} with a spoken digits database from Google Speech Commands \cite{warden2018speech}. We sampled 8,000 and 2,000 samples from the original training set as our training and validation sets. We sampled 2,000 samples from the original test set for our test set.

\textbf{LUMA} \cite{bezirganyan2024luma} includes a multimodal image-audio dataset for benchmarking multimodal learning. It contains images from a 50-class subset of CIFAR-10/100 and class label utterances. The original dataset is imbalanced, with the top 22 classes having 1,500 or more samples each. We generate a balanced set with 1,500 instances per class, totaling 33,000 pairs, which are split into 25,000 training, 4,000 validation, and 4,000 test samples.

\textbf{MM-IMDb} \cite{arevalo2017gated} is the largest publicly available multimodal dataset for movie genre multilabel classification. It contains plot summaries (text modality) and posters (image modality) for 25,959 movies. The dataset is imbalanced, with \emph{drama} being the most prevalent category (13,967 samples) and \emph{film-noir} the least prevalent (338 samples). The dataset is split into three subsets: the training, validation, and test sets contain 15,552, 2,608, and 7,799 samples, respectively.


\begin{table}[t!]
\centering
\caption{Best performance results of unimodal models on the test set for the four experimental datasets (A=Accuracy and F1=F1 score).}
\resizebox{\columnwidth}{!}{
\begin{tabular}{|c|cc|cc|}
\hline
\textbf{DATASET} & \textbf{Modality A} & \textbf{Performance} & \textbf{Modality B} & \textbf{Performance} \\ \hline
\textbf{S-MNIST} & Audio & 88.9 A. & Image & 99.0 A. \\
\textbf{CREMA-D} & Audio & 61.9 A. & Video & 60.7 A. \\
\textbf{MM-IMDb} & Text & 64.4 A. & Image & 38.9 A. \\
\textbf{LUMA} & Audio & 97.2 F1 & Image & 79.1 F1 \\
\hline
\end{tabular}}
\label{tab:unimodal_results}
\end{table}

\begin{table*}[ht!]
\centering
\caption{Accuracy results on the test set for the multimodal fusion concatenation head ($\oplus$, green column) and each modality encoder. The yellow column reports results for the overfitting modality while the blue column reports the results for the under-utilized modality. The differences in performance between the unimodal classification heads ($\Delta_{AB}$) are reported in the white column. The best results are shown in bold.}
\resizebox{\textwidth}{!}{
\begin{tabular}{|l|>{\columncolor[RGB]{230, 255, 231}}c>{\columncolor[RGB]{251, 255, 230}}c>{\columncolor[RGB]{230, 242, 255}}cc|>{\columncolor[RGB]{230, 255, 231}}c>{\columncolor[RGB]{251, 255, 230}}c>{\columncolor[RGB]{230, 242, 255}}cc|>{\columncolor[RGB]{230, 255, 231}}c>{\columncolor[RGB]{251, 255, 230}}c>{\columncolor[RGB]{230, 242, 255}}cc|>{\columncolor[RGB]{230, 255, 231}}c>{\columncolor[RGB]{251, 255, 230}}c>{\columncolor[RGB]{230, 242, 255}}cc|} 
\hline
\multirow{2}{*}{\textbf{MODEL}} & \multicolumn{4}{c|}{\textbf{CREMA-D}} & \multicolumn{4}{c|}{\textbf{S-MNIST}}  & \multicolumn{4}{c|}{\textbf{LUMA}} & \multicolumn{4}{c|}{\textbf{MM-IMDb}} \\ 

& \textbf{A$_{A \oplus V}$} & \textbf{A$_{A}$} & \textbf{A$_{V}$} & \textbf{$\Delta_{AV}$} & \textbf{A$_{I \oplus A}$} & \textbf{A$_{I}$} & \textbf{A$_{A}$} & \textbf{$\Delta_{IA}$} & \textbf{A$_{A \oplus I}$} & \textbf{A$_{A}$} & \textbf{A$_{I}$} & \textbf{$\Delta_{AI}$} & \textbf{F1 $_{T \oplus I}$} & \textbf{F1 $_{T}$} & \textbf{F1 $_{I}$} & \textbf{$\Delta_{TI}$} \\ \hline
Vanilla & 62.6 & 58.2 & 26.6 & +31.6 & 98.4 & 98.5 & 53.9 & +44.6 & 98.2 & 71.0 & 29.8 & +41.2 & 63.3 & 62.5 & 27.9 & +34.6 \\

+ MSLR-K \cite{yao2022modality} & 65.1 & \textbf{59.9} & 29.7 & +30.2 & 99.0 & \textbf{98.8} & 58.9 & +39.9 & 98.7 & 87.8 & 38.1 & +49.7 & 63.4 & 62.4 & 27.9 & +34.5\\

+ MSLR-S \cite{yao2022modality} & 64.8 & 58.2 & 29.2 & +29.0 & 99.1 & 98.6 & 66.1 & +32.5 & 98.5 & 93.2 & 35.7 & +57.5 & 63.6 & 62.7 & 28.2 & +34.5 \\

+ MSLR-D \cite{yao2022modality} & 64.3 & \textbf{59.9} & 39.7 & +20.2 & 99.0 & \textbf{98.8} & 76.9 & +21.9 & 98.7 & 93.3 & 37.0 & +56.3 & 63.6 & 62.6 & 28.5 & +34.1 \\

+ OGM \cite{peng2022balanced} & 63.9 & 57.9 & 34.6 & +23.3 & 98.9 & \textbf{98.8} &	64.5 & +34.3 & 98.5 & 86.4 & 44.0 & +42.4 & 63.7 & 63.0 & 28.3 & +34.7 \\

+ OGM-GE \cite{peng2022balanced} & 70.4 & 54.6 & 44.2 & +10.4 & 99.0 & 97.3 & 73.0 & +24.3 & 96.2 & 74.7 & 49.7 & +25.0 & 64.0 & 62.6 & 29.3 & +33.3 \\

+ MSES \cite{fujimori2020modality} & 71.6 & 57.7 & 55.9 & +1.8 & 98.4 & 98.3 & 77.8 & +20.5 & 98.1 & 88.3 & 66.2 & +22.1 & 64.4 & 63.4 & 30.6 & +32.8 \\

+ G-Blend \cite{wang2020makes} & 71.8  & 58.4 & 56.2 & +2.2 & 98.8 & 98.1 & 75.9 & +22.2 & 98.5 & 88.4 & 67.7 & +20.7 & 64.3 & 63.4 & 30.3 & +33.1 \\

\rowcolor[gray]{0.9}+ MILES (Ours) & \textbf{75.1} & \textbf{59.9} & \textbf{60.8} & \textbf{-0.9} & \textbf{99.8} & \textbf{98.8} &\textbf{84.9} & \textbf{+13.9} & \textbf{99.7} & \textbf{95.1} & \textbf{75.1} &\textbf{+20.4} &  \textbf{65.1} & \textbf{64.2} & \textbf{36.6} & \textbf{+27.6} \\ \hline
\end{tabular}}
\label{tab:concatenation}
\end{table*}

\begin{table*}[h]
\centering
\caption{Accuracy and F1 performance results on the test set for the multimodal fusion summation head ($+$, green column) and each modality encoder. The yellow column reports results for the dominant modality while the blue column reports the results for the under-utilized modality. The differences in performance between the unimodal classification heads ($\Delta_{AB}$) are reported in the white column. The best results are shown in bold.}
\resizebox{\textwidth}{!}{
\begin{tabular}{|l|>{\columncolor[RGB]{230, 255, 231}}c>{\columncolor[RGB]{251, 255, 230}}c>{\columncolor[RGB]{230, 242, 255}}cc|>{\columncolor[RGB]{230, 255, 231}}c>{\columncolor[RGB]{251, 255, 230}}c>{\columncolor[RGB]{230, 242, 255}}cc|>{\columncolor[RGB]{230, 255, 231}}c>{\columncolor[RGB]{251, 255, 230}}c>{\columncolor[RGB]{230, 242, 255}}cc|>{\columncolor[RGB]{230, 255, 231}}c>{\columncolor[RGB]{251, 255, 230}}c>{\columncolor[RGB]{230, 242, 255}}cc|} 
\hline
\multirow{2}{*}{\textbf{MODEL}} & \multicolumn{4}{c|}{\textbf{CREMA-D}} & \multicolumn{4}{c|}{\textbf{S-MNIST}} & \multicolumn{4}{c|}{\textbf{LUMA}} & \multicolumn{4}{c|}{\textbf{MM-IMDb}} \\

& \textbf{A$_{A+V}$} & \textbf{A$_{A}$} & \textbf{A$_{V}$} & \textbf{$\Delta_{AV}$} & \textbf{A$_{I+A}$}  & \textbf{Acc$_{I}$} & \textbf{A$_{A}$} & \textbf{$\Delta_{IA}$} &\textbf{A$_{A+I}$} & \textbf{A$_{A}$} & \textbf{A$_{I}$} & \textbf{$\Delta_{AI}$} &\textbf{F1$_{T+I}$} & \textbf{F1$_{T}$} & \textbf{F1$_{I}$} & \textbf{$\Delta_{TI}$} \\ \hline
Vanilla & 62.9 & 58.5 & 27.2 & +31.3 & 98.3 & 98.0 & 55.8 & +42.2 & 98.5 & 86.7 & 43.2 & +43.5 & 63.1 & 63.1 & 27.8 & +35.3 \\


+ MSLR-K \cite{yao2022modality} & 65.7 & 57.4 & 33.3 & +24.1 & 98.7 & 98.5 & 68.2 & +30.3 & 98.8 & 87.0 & 43.6 & +43.4 & 63.5 & 61.5 & 29.1 & +32.4 \\

+ MSLR-S \cite{yao2022modality} & 66.0 & 59.9 & 31.1 & +28.8 & 98.8 & \textbf{98.8} & 58.1 & +40.7 & 98.7 & 92.0 & 44.3 & +47.7 & 63.9 & 61.8 & 28.9 & +32.9 \\

+ MSLR-D \cite{yao2022modality} & 65.4 & 59.9 & 34.9 & +25.0 & 99.1 & 98.7 & 73.5 & +25.2 & 98.9 & 91.8 & 48.2 & +43.6 & 64.0 & 62.0 & 29.3 & +32.7 \\

+ OGM \cite{peng2022balanced} & 65.6 & 56.8 & 36.9 & +19.9 & 98.6 &  97.8 & 61.8 & +36.0 & 98.7 & 87.3 & 46.7 & +40.6 & 63.9 & 62.2 & 28.7 & +33.5 \\

+ OGM-GE \cite{peng2022balanced} & 68.6 & 52.5 & 43.6 & +8.9 & 99.0 & 96.9 & 79.1 & +17.8 & 97.7 & 73.2 & 48.4 & +24.8 & 63.9 & 62.0 & 29.0 & +33.0 \\

+ MSES \cite{fujimori2020modality} & 72.1 & 55.1 & 58.1 & -3.0 & 98.9 &  98.6 & 81.4 & +17.2 & 98.4 & 90.2 & 70.6 & +19.6 & 64.0 & 63.8 & 29.1 & +34.7 \\

+ G-Blend \cite{wang2020makes} & 72.3  & 54.9 & 57.6 & -2.7 & 99.2 & \textbf{98.8} & 80.1 & +18.7 & 98.7 & 90.3 & 71.1 & +19.2 & 64.3 & 63.1 & 29.1 & +34.0 \\

\rowcolor[gray]{0.9}+ MILES (Ours) & \textbf{75.7} & \textbf{60.0} & \textbf{60.8} & \textbf{-0.8} &  \textbf{99.8} & \textbf{98.8} & \textbf{85.0} & \textbf{+13.8} & \textbf{99.8} & \textbf{95.7} & \textbf{76.6} &  \textbf{+19.1} &
\textbf{65.0} & \textbf{64.0} & \textbf{33.1} & \textbf{+30.9}\\ \hline
\end{tabular}}
\label{tab:summation}
\vspace{-3.2ex}
\end{table*}

\subsection{Model training and evaluation}

\textbf{Architecture.} The \emph{vanilla} multimodal neural network architecture employed in our experiments is composed of two unimodal networks, one for each modality. We parameterize the encoders for each modality as follows: (i) BERT \cite{kenton2019bert} for the text modality, and (ii) ResNet \cite{he2016deep} variants for the image and audio modalities. The intermediate representations from the encoders are then fused and used as input for a linear layer that outputs the multimodal prediction. We evaluate all tasks using two fusion methods: concatenation and summation.

\textbf{Baselines.} We compare MILES to seven state-of-the-art approaches for balanced multimodal learning: 
\begin{itemize}
    \item MSES \cite{fujimori2020modality}  uses early stopping for balancing multimodal learning based on convergence and generalization performance.
    \item MSLR \cite{yao2022modality} uses the best unimodal learning rate to enhance modality learning. We evaluate all variants: (i) MSLR-K, (ii) MSLR-S, (iii) MSLR-D.
    \item OGM \cite{peng2022balanced} adaptively controls the optimization of each modality and OGM-GE adds noise dynamically to OGM to avoid possible generalization drops.
    \item G-Blend \cite{wang2020makes} adds multiple unimodal losses to the loss function (similar to MILES, as shown in Equation \ref{mml-loss}), and weights them based on the OGR of each modality.
\end{itemize}

\textbf{Implementation details}. We apply all of the baselines and our proposed approach on the {vanilla} architecture using the two fusion variants. We use open source implementations if available. For the CREMA-D dataset, we extract one frame from each clip and process the audio data as a spectrogram of size $257\times 299$ with window length of 512 and overlap of 353. We resize both modalities to $224\times 224$ images (3 channels for image and one for audio) and use ResNet-18 \cite{he2016deep} as backbones. We train all CREMA-D models for 200 epochs. For the S-MNIST dataset, we use the processed data provided by \cite{smnistdataset}. We reshape and resize both modalities to be $28\times28$ images (one channel) and use ResNet-10 as backbones. We train all S-MNIST models for 50 epochs. For the MM-IMDb dataset, we pre-process the data following \cite{arevalo2017gated}. We use a base BERT model with pre-trained weights, and a ResNet-50 model as text and image backbones, respectively. We train all these models for 50 epochs. For the LUMA dataset, we pre-process the data as provided by \cite{bezirganyan2024luma}, generating a mel spectrogram (96 filterbanks) for the audio modality (shape $96\times 96\times 1$) and resizing CIFAR-10/100 to $96\times 96\times 3$. We use ResNet-6 as backbones and train all models for 60 epochs. For all experiments, we use the Adam optimizer with $\beta_1 = 0.9$ and $\beta_2 = 0.999$ . We train all models using an Nvidia A100 GPU.

\textbf{Hyper-parameter tuning.} For fair comparison of methods across models, we use random hyper-parameter tuning for all experiments. We randomly sample 80 combinations of hyper-parameters per model for each dataset, fusion method, and baseline using the recommended hyper-parameter ranges suggested by authors in the original work, or based on ranges determined according to our experiments. We select the best model for each method based on the multimodal performance on the validation set and evaluate each selected model on the final test set. We report the best performance results on the test set. 

\textbf{Performance metrics.} For the S-MNIST, LUMA, and CREMA-D tasks, we report the best accuracy results on the test set, while for the highly imbalanced MM-IMDb, we report the best F1 score results on the test set. In addition, we report the difference between the unimodal classification heads for each method evaluated. Specifically, we subtract the weaker modality from the stronger modality on the best model on the test set. Formally, assuming $A$ is the stronger modality and $B$ is the weaker modality, we define it as $\Delta_{AB} = A_A(\cdot) - A_B(\cdot)$, where $A(\cdot)$ denotes the classification accuracy metric.

\textbf{Ablation studies.}
To understand the impact of each hyper-parameter in the MILES algorithm and to provide recommendations for their use, we conduct three ablation studies:

\begin{enumerate}
    \item \textbf{$\tau$ sensitivity analysis:} We vary the value of $\tau$ and analyze its impact on test set accuracy while keeping the other hyper-parameters fixed.
    \item \textbf{$\mu$ sensitivity analysis:} We vary the value of $\mu$ and analyze its impact on test set accuracy while keeping the other hyper-parameters fixed.
    \item \textbf{Computing $\mathbf{u}$ based on training metrics:} We apply MILES by computing the conditional utilization rates (Eq. \ref{eq:utilization}) using the training set performance metrics instead of the validation set metrics.
\end{enumerate}

For the sensitivity analysis studies, we fix the learning rate to the one corresponding to the best results. For the third ablation study, we perform hyper-parameter tuning as described in the previous paragraph.

\section{Results}
This section provides the experimental results for unimodal baselines, multimodal performance and all ablation studies. Based on these, we derive some recommendations on how to use MILES more efficiently.

\subsection{Unimodal performance results}

We report the results of the best unimodal models for each test set in Table \ref{tab:unimodal_results} for comparison with multimodal model performance. As observed, the four datasets exhibit varying performance across modalities. Specifically, for CREMA-D, despite being a more complex task (emotion recognition) than the other three tasks, both modalities have similar unimodal accuracy (61.9\% vs. 60.7\%). The S-MNIST dataset is relatively easier for unimodal models, with image data achieving 99\% accuracy and 88.9\% accuracy for the audio modality. For the LUMA dataset, the audio modality outperforms the visual modality by a wide margin (97.2\% vs. 79.1\%). Finally, for the MM-IMDb task, multilabel movie genre classification, the text modality is significantly better than the image modality (64.4\% vs. 38.9\%). 

\subsection{Multimodal performance results}

\textbf{Concatenation}. Table \ref{tab:concatenation}
shows the results for all baseline models and MILES on the four datasets using feature concatenation as the fusion method. MILES consistently outperforms all baselines on multimodal and unimodal performance across datasets, with the greatest margin for the CREMA-D dataset, which has the lower baseline performance (62.6\% vs. 75.1\% for multimodal performance). Apart from enhancing multimodal prediction, MILES produces strong modality encoders with accuracy close to the unimodal baselines shown in Table \ref{tab:unimodal_results}. The modality difference measure, $\Delta_{AB}$, reaches its minimum value for MILES in all tasks, indicating that MILES can effectively balance modality usage for both multimodal and unimodal predictions. More interestingly, for the CREMA-D task, the difference is negative, indicating that the weaker modality became the stronger for the MILES model, with the stronger modality still on par with the best baselines methods. Comparing the unimodal models (see Table \ref{tab:unimodal_results}) and unimodal encoders (see Table \ref{tab:concatenation}), on average, there is a gap of $\approx 1.8 \pm 1.6\%$ in performance between the best unimodal model and the top-performing MILES modality encoders (across the eight encoders), with the smallest gap being $0\%$ for CREMA-D video and the largest gap being $4\%$ for the LUMA image encoder. In all cases, MILES boosts the predictive performance of the non-dominant modality, especially for LUMA (29.8\%  vs. 75.1\%), while keeping top accuracy metrics for the dominant modality. 

\textbf{Summation}. Table \ref{tab:summation} shows the results for all baseline models and MILES on the four datasets using feature summation as the fusion method. Similar to the setting using feature concatenation, MILES outperforms all baselines on multimodal and unimodal performance across all datasets. While we observe the same overall trends as for concatenation, the results for the feature summation technique are greater than for concatenation for most models. The same observations as in the concatenation setting apply for the difference between encoders, $\Delta_{AB}$, with MILES providing the minimum values, thereby producing not only stronger but also more balanced encoders. Comparing the performance of unimodal models with the unimodal encoders of the multimodal networks, MILES consistently produces the best multimodal results across all datasets and also generates the strongest unimodal encoders, with an average performance gap between the best models and MILES encoders of $\approx 2.0  \pm  2.0\%$ (ranging from $0\%$ for the CREMA-D video encoder to 5.8\% for the MM-IMDb image encoder).

Overall, these results show that regardless of the fusion method employed and across datasets and diverse backbone architectures, MILES allows overcoming the modality overfit of conventional training, surpassing all state-of-the-art methods in enhancing both multimodal and unimodal performance, especially for the non-dominant modality. 

\subsection{Ablation studies results}

In the following paragraphs, we provide the results for the ablation studies, which were performed using the LUMA dataset. For both hyper-parameter sensitivity analyses, we use the best configurations for MILES in our experiments ($\tau$ = 0.2 and $\mu$ = 0.5).

\textbf{$\tau$ sensitivity analysis}. Table \ref{tab:tau_sensitivity} shows the results of varying $\tau$ while keeping the other hyper-parameters fixed (using the hyper-parameters of the best MILES model on the LUMA dataset from Table \ref{tab:concatenation} and Table \ref{tab:summation}). The results show that smaller $\tau$ values (e.g., $\tau$ = 0.0) force the model to balance both modalities, boosting the performance of the non-dominant modality. However, this comes at the cost of notably lowering the performance of the dominant modality, which negatively impacts both unimodal and multimodal performance, resulting in improved but sub-optimal models.

The same occurs at the other end of the spectrum when $\tau$ is larger (e.g., $\tau$ = 0.5), as it does not restrict the learning of the dominant modality, allowing the model to overfit. Nevertheless, the multimodal model improves the non-dominant modality performance via the added modality-specific losses (Equation \ref{mml-loss}), but the results are suboptimal with respect to the best MILES results.

\begin{table}[t]
\centering
\caption{Results for $\tau$ sensitivity analysis. We vary $\tau$ while keeping the learning rate and $\mu$ hyper-parameters fixed. The best results are shown in bold.}
\resizebox{0.74\columnwidth}{!}{
\begin{tabular}{|c|c|>{\columncolor[RGB]{230, 255, 231}}c>{\columncolor[RGB]{251, 255, 230}}c>{\columncolor[RGB]{230, 242, 255}}c|}
\hline
\textbf{$\tau$} & \textbf{Fusion} & \textbf{Acc$_{IA}$} & \textbf{Acc$_{A}$} & \textbf{Acc$_{I}$} \\ \hline
0.5 & \multirow{5}{*}{ $\oplus$ } &  98.9 & 97.4 & 57.6  \\ 
0.3 & & 99.1 & 95.4 & 66.7  \\ 
0.2 & & \textbf{99.7} & \textbf{95.1} & \textbf{75.1} \\ 
0.1 & & 99.1 & 93.1 & 76.8  \\  
0.0 & & 98.7 & 91.7 & 77.5  \\ \hline

0.5 & \multirow{5}{*}{ $+$ } & 99.1 & 97.3 & 66.6  \\ 
0.3 & & 99.2 & 95.8 & 73.2  \\ 
0.2 & & \textbf{99.8} & \textbf{95.7} & \textbf{76.6}  \\ 
0.1 & & 99.3 & 93.3 & 77.4  \\ 
0.0 & & 98.9 & 91.5 & 78.1 \\ \hline
\end{tabular}}
\label{tab:tau_sensitivity}
\end{table}

\textbf{$\mu$ sensitivity analysis}. Table \ref{tab:mu_sensitivity} reports the results of varying $\mu$ while keeping the other parameters fixed using the hyper-parameters of the best MILES model on the LUMA dataset (Table \ref{tab:concatenation} and \ref{tab:summation}). The results show that smaller values of $\mu$, such as 0.05 and 0.01, which imply a greater reduction in the learning rate of the dominant modality, significantly affect the learning of that specific modality. This impacts overall model performance, enhancing the non-dominant modality's performance but potentially degrading the dominant modality's performance. On the contrary, larger values closer to one may not enhance the learning of the non-dominant modality. The value of $\mu = 1$ is equivalent to greater values of $\tau$ (e.g., $\tau = 0.5$), disabling the effects of MILES on the learning process. 

\begin{table}[t]
\centering
\caption{Results for $\mu$ sensitivity analysis. We vary $\mu$ while keeping the learning rate and $\tau$ hyper-parameters fixed. The best results are shown in bold.}
\resizebox{0.74\columnwidth}{!}{
\begin{tabular}{|c|c|>{\columncolor[RGB]{230, 255, 231}}c>{\columncolor[RGB]{251, 255, 230}}c>{\columncolor[RGB]{230, 242, 255}}c|}
\hline
\textbf{$\mu$} & \textbf{Fusion} & \textbf{Acc$_{IA}$} & \textbf{Acc$_{I}$} & \textbf{Acc$_{A}$} \\ \hline
1.00 & \multirow{8}{*}{ $\oplus$ } & 98.7  & 97.5 & 57.7 \\ 
0.75 & & 99.4 & 96.9 & 65.5  \\
0.50 & & \textbf{99.7} & \textbf{95.1} & \textbf{75.1}  \\ 
0.25 & & 99.1 & 94.5 & 75.4   \\ 
0.10 & & 94.7 & 92.2 & 76.8  \\ 
0.05 & & 90.3 & 89.3 & 77.7   \\
0.01 & & 86.7 & 84.4 & 79.0  \\ \hline
1.00 & \multirow{8}{*}{ $+$ } & 99.0 & 97.3 & 66.7 \\ 
0.75 & &  99.7 & 96.7 & 69.3  \\ 
0.50 & & \textbf{99.8} & \textbf{95.7} & \textbf{76.6}  \\ 
0.25 & &  99.4 & 94.9 & 76.8  \\ 
0.10 & &  97.6 & 93.9 & 78.2  \\ 
0.05 & &  91.5 & 88.8 & 78.9  \\
0.01 & &  89.6 & 83.9 & 79.4  \\  \hline
\end{tabular}}
\label{tab:mu_sensitivity}
\end{table}

\textbf{Computing $\mathbf{u}$ based on training metrics}. Table \ref{tab:training_metrics} presents the results of comparing training versus validation metrics for calculating the conditional utilization rate, which is a crucial epoch-wise computation in the MILES algorithm, as outlined in Algorithm \ref{alg:mils}. The results indicate that using training metrics for conditional utilization rate computation can yield comparable outcomes. However, validation metrics generally achieve the best results more quickly during the training process compared to training set metrics. In both cases, the training metrics produced the best model at epoch 53 out of a total of 60 epochs, while the best models for validation occurred slightly earlier, at epochs 45 and 47. This demonstrates the versatility of the MILES algorithm and its potential application in tasks where a proper validation set is unavailable. We note that Table \ref{tab:concatenation} and Table \ref{tab:summation} report results of models that track the validation set metrics, for consistency and fair comparison with other methods that require the availability of a validation set (e.g., MSES).

\begin{figure*}[t]
\centering
\includegraphics[width=\textwidth]{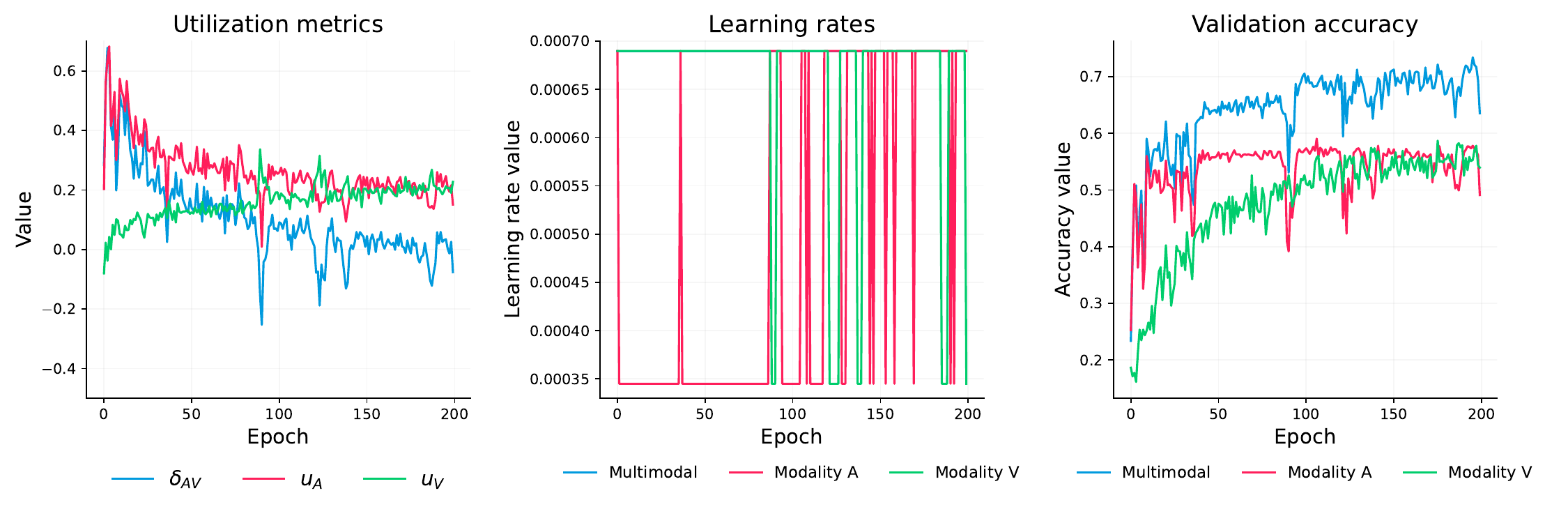}
\caption{Epoch-wise utilization metrics, learning rates, and validation accuracy for an example run on the CREMA-D dataset allow for the visualization of multimodal and unimodal learning dynamics using MILES.}
\label{fig:learning_dynamics}
\end{figure*}

\begin{table}[t!]
\centering
\caption{Comparison of MILES results using training vs. validation set metrics for conditional utilization rate calculation. The best results are shown in bold.}
\resizebox{\columnwidth}{!}{
\begin{tabular}{|c|c|>{\columncolor[RGB]{230, 255, 231}}c>{\columncolor[RGB]{251, 255, 230}}c>{\columncolor[RGB]{230, 242, 255}}c|c|}
\hline
\textbf{Fusion} & \textbf{$\mathbf{u}$} & \textbf{Acc$_{IA}$} & \textbf{Acc$_{A}$} & \textbf{Acc$_{I}$} & \textbf{Epoch} \\ \hline
\multirow{2}{*}{ $\oplus$ } & Training  & \textbf{99.8} & \textbf{95.6} & \textbf{73.6}  & 53 \\ 
 & Validation & 99.7 & 95.1 & 75.1 & \textbf{47}  \\ \hline
\multirow{2}{*}{ $+$ } & Training  & \textbf{99.8}  & \textbf{97.3} & 76.4 & 53 \\ 
 & Validation & \textbf{99.8} & 95.7 & \textbf{76.6} & \textbf{45} \\ \hline
\end{tabular}}
\label{tab:training_metrics}
\end{table}

\textbf{Recommendations}. While the impact and benefit of MILES on multimodal training may vary across datasets, fusion methods, architectures, and input modalities, as shown in Table \ref{tab:concatenation} and Table \ref{tab:summation}, our empirical experience and ablation study results, enable us to provide a few recommendation for the effective usage of MILES that may work well across settings:

\begin{itemize}
    \item Training of unimodal models can provide a good reference on the starting point for $\tau$ and $\mu$ hyper-parameters. For example, if unimodal models are very close in performance, the multimodal model may be able to integrate both modalities naturally in a more balanced way. Thus, $\tau$ could likely be set to smaller values (e.g., $\tau = 0.0$), whereas if the performance gap is significant, setting $\tau = 0.0$ may not lead to good outcomes.
    \item In general, start experiments with a moderate value like $\tau = 0.2$ and increase or decrease according to results.
    \item Initially, set $\mu$ based on the ratio between the best learning rates of unimodal models. Then, adjust $\mu$ by decreasing or increasing its value as needed. Smaller values of $\mu$ work better, as shown in Table \ref{tab:mu_sensitivity}, because they do not excessively hinder the learning of the dominant modality.
    \item Once a sensitive range of values is identified for both hyper-parameters, fine-grained hyper-parameter tuning is recommended. However, in many cases, it may be sufficient to keep them fixed and vary only the learning rate.
\end{itemize}

\section{Learning dynamics}

To visualize the multimodal and unimodal learning dynamics during training using MILES, Fig.~\ref{fig:learning_dynamics} shows the epoch-wise values of utilization metrics, learning rates, and validation accuracy for an example training run on the CREMA-D dataset. Note that this is not the best run; it is provided solely for the purpose of visualizing the learning dynamics using MILES, as defined in Algorithm~\ref{alg:mils}. In this example, the hyper-parameters were: (i) learning rate $= 0.00068983$, (ii) $\tau = 0.05$, and (iii) $\mu = 0.5$.

As can be seen in Fig.~\ref{fig:learning_dynamics}, the utilization of the weaker modality improves during training, with the value of $\delta_{AV} \approx 0$ (equal utilization of the modalities) at the end of the training. Throughout the training process, the learning rates are dynamically adjusted every epoch following Algorithm \ref{alg:mils} and based on the pre-specified $\tau$ and $\mu$ values. Only one of the modality-specific learning rates is adjusted, while the multimodal learning rate is kept fixed. This allows both the utilization metric and validation accuracy to improve over time until the end of the training process.

\section{Limitations}
Seminal work on multimodal learning focus on the bimodal scenario of multimodal learning \cite{wu2022characterizing, huang2022modality, wang2020makes}. Similarly, all state-of-the-art baseline approaches considered in our work use two modalities to show and demonstrate their improvements \cite{peng2022balanced, yao2022modality, fujimori2020modality}. Following this standard, we showcase MILES for the bimodal case, which is the most frequent setting in current state-of-the-art research related to multimodal learning \cite{xu2023multimodal}. In addition, most benchmark datasets are bimodal. However, MILES could be effectively adapted to scenarios with more than two modalities. For instance, for a third modality C, Algorithm \ref{alg:mils} would need to incorporate the computation of the conditional utilization rate of the third modality ($\mathbf{u_C}$), calculate pairwise differences for all modalities ($\delta_{AB}$, $\delta_{BC}$, $\delta_{AC}$) and include a conditional block (lines 8-21 in Algorithm \ref{alg:mils}, adaptation may be required) for each difference, maintaining the same assumptions. Notwithstanding that, future work will focus on assessing the validity of these assumptions for more than two modalities and their interplay.

During our experiments, we noticed that, depending on the initialization of the network parameters, MILES might be excessively penalizing the learning of the dominant modality. This may hinder its effective learning and result in sub-optimal performance models for the dominant modality, thus requiring additional rounds of hyper-parameter tuning to achieve optimal learning. We aim to address this caveat in future work to make the proposed framework more efficient.

\section{Discussion and Future Work}

The potential of multimodal machine learning to enhance unimodal performance is often hindered by training challenges such as modality overfitting. In this work, we propose MILES, a learning rate scheduler designed to balance and enhance multimodal machine learning. MILES leverages the epoch-wise conditional utilization rate (using validation or training performance metrics) to balance multimodal learning during training, adjusting the learning speed of the modalities to avoid both overfitting of the dominant modality and promote utilization of the non-dominant modality. Our results show that MILES effectively addresses this challenge, outperforming seven state-of-the-art baselines across datasets, tasks, fusion methods and modalities, improving the capabilities of vanilla multimodal fusion architectures for both multimodal and unimodal predictions. In addition, MILES training produces strong unimodal encoders, enabling the use of their predictive capabilities when dealing with samples that have missing modalities. MILES is governed by two hyper-parameters, which can be tuned to effectively emphasize modality learning. We provide general recommendations on how to tune these hyper-parameters to effectively balance multimodal learning during training.  

Most multimodal machine learning methods and datasets focus primarily on the bimodal setting. However, as the field rapidly advances, datasets incorporating more modalities are slowly becoming available. Future work will explore extending our approach to these more complex scenarios and how the current assumptions expand to those cases, investigating the interplay between multiple modalities, and how to effectively address the training challenges of these networks.

\section{Acknowledgment}

This work was supported by the NYUAD Center for Interacting Urban Networks (CITIES), funded by Tamkeen under the NYUAD Research Institute Award CG001, the Center for Cyber Security (CCS), funded by Tamkeen under NYUAD RRC Grant No. G1104, and the NYUAD Center for Artificial Intelligence and Robotics, funded by Tamkeen under the NYUAD Research Institute Award CG010.  The research was carried out on the High Performance Computing resources at New York University Abu Dhabi.

\bibliographystyle{IEEEtran}
\bibliography{IJCNN/IEEEabrv, IJCNN/bibliography}

\end{document}